\documentclass[fleqn,10pt]{wlscirep}
\usepackage[utf8]{inputenc}
\usepackage[T1]{fontenc}
\usepackage{amsmath,amssymb,amsthm,mathrsfs,amsfonts,dsfont,amscd,keyval}
\usepackage{mathtools,mathrsfs}
\usepackage{yfonts} 
\usepackage{subfigure, epsfig,float}
\usepackage{braket}
\usepackage{bm}
\usepackage{fancyhdr}
\usepackage{enumitem}
\usepackage{color}
\usepackage{booktabs}
\usepackage{hyperref}
\usepackage{tabularx}
\usepackage{times}
\usepackage{multirow}
\usepackage{tikz-cd}
\usepackage{tikz-feynman}
\usepackage{capt-of}
\usepackage{tensor}
\usepackage{algorithm2e}
\usepackage{paralist}
\usepackage{algorithmic}
\usepackage{blindtext}
\usepackage{setspace}
\usepackage{float}
\usepackage{pdfpages}
\usepackage{amsmath} 
\usepackage{amsfonts} 
\usepackage{amssymb} 
\usepackage{soul} 

\renewcommand\vec{\mathbf}

\theoremstyle{definition}

\theoremstyle{plain}

\theoremstyle{remark}

\newcommand{\grp}{\mathbb{G}}   

\newcommand{\otimescg}{\otimes_{\rm cg}}

\newcommand{\comments}[1]{}

\newcommand{\risi}[1]{{}}
\newcommand{\risic}[1]{{}}
\newcommand{\risist}[1]{}

\definecolor{color-Han}{rgb}{0.6,0.1,0.1}
\definecolor{dark-green}{rgb}{0.1,0.6,0.1}

\newlength{\sectionpadding}
\newlength{\figpadding}
\newlength{\subsectionpadding}
\begin{document}

\title{Unifying O(3) Equivariant Neural Networks Design with Tensor-Network Formalism }

\author[1,*]{Zimu Li}
\author[2,*]{Zihan Pengmei}
\author[4,*]{Han Zheng}
\author[3]{Erik Thiede}
\author[5,6,7,+]{Junyu Liu}
\author[4]{Risi Kondor}

\affil[1]{DAMTP, Center for Mathematical Sciences, University of Cambridge, Cambridge CB30WA, UK}
\affil[2]{Department of Chemistry, The University of Chicago, Chicago, IL 60637, USA}
\affil[3]{Center for Computational Mathematics, Flatiron Institute, New York NY 10010}
\affil[4]{Department of Computer Science, The University of Chicago, Chicago, IL 60637, USA}
\affil[5]{Pritzker School of Molecular Engineering, The University of Chicago, Chicago, IL 60637, USA}
\affil[6]{Kadanoff Center for Theoretical Physics, The University of Chicago, Chicago, IL 60637, USA }
\affil[7] {SeQure, Chicago, IL 60615, USA}

\affil[+]{\url{junyuliu@uchicago.edu}}
\affil[*]{Contributed equally}


\begin{abstract}

Many learning tasks, including learning potential energy surfaces from ab initio calculations, involve global spatial symmetries and permutational symmetry between atoms or general particles.  Equivariant graph neural networks are a standard approach to such problems, with one of the most successful methods employing tensor products between various tensors that transform under the spatial group. However, as the number of different tensors and the complexity of relationships between them increase, maintaining parsimony and equivariance becomes increasingly challenging. In this paper, we propose using fusion diagrams, a technique widely employed in simulating SU($2$)-symmetric quantum many-body problems, to design new spatial equivariant components for neural networks. This results in a diagrammatic approach to constructing novel neural network architectures. When applied to particles within a given local neighborhood, the resulting components, which we term "fusion blocks," serve as universal approximators of any continuous equivariant function defined on the neighborhood. We incorporate a fusion block into pre-existing equivariant architectures (Cormorant and MACE), leading to improved performance with fewer parameters on a range of challenging chemical problems. Furthermore, we apply group-equivariant neural networks to study non-adiabatic molecular dynamics of stilbene cis-trans isomerization. Our approach, which combines tensor networks with equivariant neural networks, suggests a potentially fruitful direction for designing more expressive equivariant neural networks.
\end{abstract}


\flushbottom
\maketitle
%
%
\thispagestyle{empty}

\section*{Introduction}
Graph neural networks (GNNs) have recently gained prominence in the field of chemistry, owing to their ability to learn from the structural properties of molecules and materials. Nevertheless, devising an efficient and accurate GNN architecture for investigating dynamic properties of chemical systems remains a formidable challenge. GNNs are adept at learning the structure of chemical systems and predicting their properties, including potential energy, dipole moment, and atomic forces. Recently, there has been a surge of interest in employing deep learning to forecast chemical properties and expedite first-principles dynamics simulations \cite{wang2018deepmd,schoenholz2020jax,atz2021geometric,walters2020applications,shuaibi2021rotation}. Specifically, GNNs have been utilized to estimate the potential energy with distinct atomic coordinates, where the negative gradient concerning the input coordinates naturally corresponds to the atomic force. Accurate prediction of potential energy and atomic force\cite{Chmiela2016a} necessitates adherence to spatial symmetries, such as translational and rotational covariance, since these properties are continuous functions defined on three-dimensional Euclidean space.

Machine learning algorithms employed to predict properties such as potential energy and atomic forces must yield consistent results, regardless of the molecule's rotational pose or ordering. To address this challenge, researchers have developed group-equivariant neural networks that preserve these symmetries\cite{esteves2020theoretical,cohen2021equivariant,shuaibi2021rotation,Welling2021EGNN,Du2021SE3}. In a group-equivariant network, symmetry operations on the data, including rotations of pictures and molecules, and permutations of the labels of each particle, commute with the network's layers, ensuring that the same physical property is predicted irrespective of the input's orientation. Many state-of-the-art spatially equivariant neural networks\cite{batzner2021e3equivariant, batatia2022mace} leverage the representation theory of the spatial rotation group in the so-called Fourier space \cite{Kondor_clebschgordan2018}. These Fourier space methods employ the Clebsch-Gordan nonlinearities\cite{Kondor_SO3_2018, Kondor_clebschgordan2018}. In fact, as elucidated in the supplementary material (SM), Clebsch-Gordan nonlinearities are the sole source of nonlinearity in Fourier space determined by invariant theory in mathematics \cite{Olver1999,Procesi2007,Goodman2009}. The Clebsch-Gordan nonlinearities further constrain the use of linear weights, which can only act on the multiplicity space corresponding to each irreducible representation\cite{Kondor_SO3_2018}.

Independently of work in machine learning, physicists have been using network models, called \emph{tensor networks}, to represent complicated quantum many-body systems. Tensor networks are a family of methods for approximating larger tensors by contracting together a large collection of smaller tensors. Tensor networks have been used to successfully approximate large quantum states with low entanglement accurately by making use of the density matrix renormalization group (DMRG) in one dimension~\cite{white1992density,rommer1997class} and introducing low-rank tensors to represent the quantum states~\cite{Orus_2019}. Applications of tensor networks include quantum simulation of quantum physics problems~\cite{fannes1992finitely,verstraete2004renormalization,vidal2007entanglement}, quantum computing and quantum supremacy experiments~\cite{huang2021efficient,pan2021simulating}, machine learning and data science~\cite{stoudenmire2016supervised,efthymiou2019tensornetwork,roberts2019tensornetwork}, and quantum gravity~\cite{swingle2012entanglement,pastawski2015holographic}. A special type of tensor networks concerns with global on-site SU($2$) symmetry, called the \emph{spin networks}, where multiple sites are fused by a prescribed fusion diagrams~\cite{Makinen2019,Sukhwinder2012} so that the global wavefunctions are SU($2$) symmetric. The fusion diagrams, as we will show later and further in SM, are natural and sparse generalization of Clebsch-Gordan products among multiple sites. Fusion diagrams have found great success in simulate SU($2$)-symmetric quantum systems~\cite{Orus_2019,Sukhwinder2012,Schmoll2020,Schmoll2020b}, and we will show their potential for constructing universal equivariant neural networks. 

Fusion diagrams facilitate the classification of existing neural network architectures and inspire the development of novel equivariant blocks. We showcase the computational power of these blocks using classical results from invariant theory, which establish that under certain conditions, they can achieve universality. For instance, we employ fusion diagrams to construct a new SO(3)-equivariant block, which we incorporate into two state-of-the-art neural network architectures: Cormorant~\cite{Kondor_cormorant2019} and MACE~\cite{batatia2022mace}. We demonstrate that integrating the new equivariant layer significantly enhances the performance of both architectures, with a comparable or substantially fewer number of parameters.

To assess the validity of the fusion block, we carried out extensive experiments on various chemical systems, including standard benchmark datasets such as QM-9\cite{ramakrishnan2014quantum} and MD-17\cite{Chmiela2016a}, which aims to predict the quantum properties of molecules and potential energy surfaces, as well as more challenging systems like the non-adiabatic cis-trans isomerization of stilbene. Non-adiabatic isomerization of stilbene poses a considerable challenge learning the multiple boarder and reactive potential energy surfaces (PESs), necessitating accurate interpolation and extrapolation.

In summary, this paper presents a novel method for constructing group-equivariant neural network blocks using fusion diagrams, a concept borrowed from theoretical physics. Our approach alleviates the combinatorial complexity associated with preserving symmetry constraints in neural networks, enabling the construction of expressive and universal equivariant layers. We demonstrate the effectiveness of the fusion block by incorporating our new SO(3)-equivariant layer into two state-of-the-art molecular neural network architectures, Cormorant and MACE, and evaluating them on a variety of common benchmarks in companion with more complicated molecular isomerization and adsorption process. Our results indicate that the fusion block leads to improved performance with comparable or fewer parameters. Overall, our approach contributes to the developing a new routine that can be used to construct more expressive group equivariant neural networks.

\section*{Background}

Before delving into the specifics of our approach, it is crucial to lay the groundwork with some foundational concepts. In this section, we offer an overview of relevant ideas from both machine learning and physics. We begin with a concise review of molecular dynamics and the significance of symmetry and equivariance in machine learning. Subsequently, we introduce the concept of tensor products and their role in theoretical physics, including a description of the fusion diagram notation. 

\subsection*{Molecular dynamics}

Molecular dynamics simulations are essential tools for studying molecular properties at the atomic level within specific timescales. To simulate atomic motion, we need to calculate the potential energy and atomic forces acting on molecules with particular geometric configurations in $\mathbb{R}^3$ space. Generally, potential energy and its gradients can be accurately determined by electronic structure calculations from first principles or approximated classically as simple analytical potential functions within specific chemical environments, such as atomic type, bond length, and bond angle. However, the electronic structure calculations under the ab initio molecular dynamics (AIMD) calculations are expensive. 

One popular approach to overcoming this limitation is to use neural networks as interatomic potentials \cite{schoenholz2020jax,wang2018deepmd}, which are trained with reference AIMD data. Training neural networks as interatomic potentials involves regressing on potential energy and atomic forces simultaneously, where predictive forces can be naturally achieved as the negative gradient of energy via back-propagation. The 
potential energy is invariant to 3D rigid rotations, while atomic forces are covariant to rotations, as they are vector values. The equivariant neural networks that we introduce subsequently are a powerful data-driven approach for an accurate representation of the chemical environment.


\subsection*{Representation theory of SU(2) and SO(3)}

Rotationally equivariant nets are arguably one of the most successful types of equivariant neural networks. Let $X$ and $Y$ be the input and output spaces of a layer $L$, and let $T$ and $T'$ be linear actions of a group $\grp$ encoding the symmetry on $X$ resp.~$Y$. The layer is said to be equivariant to $\grp$ if 
\begin{equation}
    {T}_g' \circ L = L \circ T_g \qquad \text{for all } g \in \grp.
    \label{eq:defn_equiviariance}
\end{equation}
If the group action on the output space is the identity transformation, i.e. ${T}_g' y = y$ for all elements of $\grp$, the above reduces to
\begin{equation}
    L = L \circ T_g \qquad \forall g \in \grp
    \label{eq:defn_invariance}
\end{equation}
and we have an \emph{invariant} layer. Constructing an equivariant neural network requires that both the learned affine function and the fixed nonlinear function obey equivariance. Kondor et al.\cite{Kondor_SO3_2018} showed how to construct learned affine functions that are equivariant to compact groups (such as the group of rotations or the group of permutations) using the theory of linear representations. A linear representation \footnote{Linear representations should not be confused with the different use of the word ``representation'' in representation learning.} of a compact group $G$ is pair $(V, \rho)$ such that for each $g \in G$, $\rho(g)$ is assigned a linear transformation of $V$ for which $\forall g_1, g_2 \in G , \rho(g_1) \rho (g_2) = \rho(g_1 g_2)$. If the representation is finite-dimensional, the range of $\rho$ is a subset of the space of complex $k\, \times \, k$ matrices for some $k$.  An irreducible linear representation (irrep) is a representation where $V$ has no proper subspaces preserved under $\rho$.
Using well-known results in representation theory, we can apply a linear transformation that decomposes the inputs, outputs, and activations of a neural network into components that transform according to the group's irreps.
Then, one can show that the most general possible equivariant linear transformation can be written as matrix multiplication against each component \cite{Zaheer2017,Kondor_SO3_2018,Segol2019,Maron2020sym}.
The construction of linear equivariant layers where inputs and outputs transform according to linear group representations has been widely studied and used in today's neural networks \cite{Cohen2016,Zaheer2017,Kondor_SO3_2018,Segol2019,Maron2019,Kondor_Sn2020,Welling2021EGNN,Du2021SE3}. 

For the rest of this work, we will focus on equivariance in the presence of SU(2) and SO(3) symmetries.  These groups have fundamental importance in modern quantum physics and machine learning applications on geometric data. The irreps of SU(2) can be indexed by a single non-negative integer or half-integer, called the spin label. For any $g \in\text{SU(2)}$ and spin label $j$, we denote the corresponding matrix that arises from evaluating $\rho(g)$ as  $W^{j}(g)$.
It is well known in group theory that SO(3) irreps are isomorphic to the irreps of SU(2) with integer spin labels \cite{Varshalovich1988,lindgren1986,Goodman2009}. This relationship allows us to study both SO(3) and SU(2) at the same time. Depending on the mathematical context, vectors in $V_j$ might transform either by $W^{j}(g)$ (contravariant transformation) or by its complex conjugate (covariant transformation). 
In what follows we focus only on irreps and omit $\rho_j$ when we denote irreps. 
To distinguish these two cases, we denote components of any vector $\vec{v}$ transforming contravariantly by raised index $v^m$ with $w_m$ being defined accordingly for the covariant case. With the notion of raised and lowered indices, one can contract vectors like $v^m w_m$, where the Einstein summation convention will be used consistently in this paper. 

With the above basic notions clarified, let us formally define the Clebsch-Gordan product.
Take two SU(2) irreps $(\rho_{j_a}, V_{j_a})$ and $(\rho_{j_b}, V_{j_b})$
of spin $j_a$ and $j_b$ respectively. 
We can then define the tensor product representation $(\rho_{j_a} \otimes \rho_{j_b}, V_{j_a} \otimes V_{j_b})$. As this is still an SU(2) representation, it can be decomposed into irreps labeled by spins. A \emph{Clebsch-Gordan decomposition} is a matrix $C^{(j_a,j_b,j_c)}$ which transforms the tensor product $W^{j_a}(g) \otimes W^{j_b}(g)$ into $W^{j_c}(g)$ for a prescribed spin $j_c$ and \emph{any} $g \in$ SU(2). 
Formally,
\begin{align}
	C^{(j_a,j_b,j_c) \dagger} \big( W^{j_a}(g) \otimes W^{j_b}(g) \big) C^{(j_a,j_b,j_c)} =  W^{j_c}(g)  \ \text{ for all } g \in \text{SU}(2).
\end{align} 
By definition, the Clebsch-Gordan decomposition is equivariant with respect to the action of SU(2) as well as SO(3). 
Formally, it can be understood an element from the space of SU(2) equivariant maps $\operatorname{Hom}(V_{j_a} \otimes V_{j_b}, V_{j_c})$, where $V_{j_a}, V_{j_b}, V_{j_c}$ are the corresponding $\mathrm{SO}(3)$ irreps with the angular momenta labels $j_1, j_2, j_3$. In this case, we can write the Clebsch-Gordan product as a third-order tensor: 
\begin{align}
C^{(j_a, j_b, j_c)}{}_{m_1, m_2}{}^{m_3} \in \operatorname{Hom}(V_{j_a} \otimes V_{j_b}, V_{j_c}),
\end{align}
where $m_1, m_2, m_3$ are the corresponding magnetic quantum numbers. There are well-established methods to compute $C^{(j_a,j_b,j_c)}$ both theoretically and algorithmically, e.g., \cite{BiedenharnUnWigner1967,Chen1985, Varshalovich1988}. Summing in Einstein notation with lower and upper indices, we call it a Clebsch-Gordan product for input vectors $\psi^{(j_a)}, \psi^{(j_b)}$:
\begin{align} \label{cg-product}
	\psi^{(j_c)}{}^{m_c} = C^{(j_a, j_b, j_c)}{}_{m_a, m_b}{}^{m_c} \psi^{(j_a)}{}^{m_a} \psi^{(j_b)}{}^{m_b}.
\end{align}
We will also leave $m$ as entry indices and write Clebsch-Gordan product among input vectors as inner products:
\begin{align} \label{cg-projection}
	\psi^{(j_c)} = \langle C^{(j_a, j_b, j_c)}, \psi^{(j_a)} \otimes \psi^{(j_b)} \rangle
\end{align}

\section*{Methods}

Here we demonstrate how fusion diagrams can be used to design equivariant components that we call ``neural fusion blocks.'' We present an explicit construction for transformations under SU(2) and SO(3).In each block, we apply a collection of fusion diagrams $\mathcal{Q}$ to our input tensors. Each incoming edge of the diagram is associated with an input to the block and each outgoing edge is associated with an output of the block.
    We denote the collection of input tensors associated with incoming edges as $\{ \Psi^{(\operatorname{in})}_1,...,\Psi^{(\operatorname{in})}_n\}$ and components
    corresponding to the spin label $j_1,...j_n$ are denoted as $\{\psi^{(\operatorname{in}, j_1)}_{\tau_1},...,\psi^{(\operatorname{in}, j_n)}_{\tau_n} \}$, where 
    $\tau_1,...,\tau_n$ are the channel dimension indices.
    We omit batch indices from our treatment for brevity.
    The fusion block then acts according to Algorithm~\ref{alg:fusion-block}. More illustrations on the updating rule with diagrammatic examples can be found in Section I in the SM.
    Due to the use of fusion diagrams, the resulting algorithm is guaranteed to be equivariant to rotations.It also serves an rotationally-equivariant universal approximator where we put the proof details in Section II in the SM.
    
    \begin{algorithm}[tb!]
       \caption{Fusion Block \risic{In general, an "algorithm" figure contains pseudocode and less text.}}
       \label{alg:fusion-block}
    \begin{algorithmic}
       \STATE {\bfseries Input:} Incoming atomic activations: $\{\Psi^{(\operatorname{in})}_1, \cdots \Psi^{(\operatorname{in})}_n\}$ and to initialize a $\operatorname{FusionBlock}$ module
        \STATE{\bfseries Output:} Outgoing atomic activations:  $ \Psi^{(\operatorname{out})}$
       \STATE 
       \FOR{$Q^{(\vec{j}, \vec{k}; J)}_i \in \mathcal{Q}$}
       \STATE apply incoming activations$\psi_1^{(\operatorname{in},j_1)}, \ldots \psi_n^{(\operatorname{in}, j_n)}$.
    		with an aggregation function $\phi$ to enforce permutation equivariance:
    		\begin{align}
    			\widetilde{\psi}_i^{J} = \phi(\operatorname{FusionBlock}\left( \psi^{(\operatorname{in}, j_1)}_1 \cdots \psi^{(\operatorname{in}, j_n)}_n \right))
    		\end{align}
       \ENDFOR
       \STATE Concatenate outputs for each fusion diagram along the new channel dimension$\tilde{\tau}_c$ and reshape. 
    	\begin{align}
    		\tilde{\Psi}_{\tilde{\tau_c}} = \tilde{\Psi}_1 \oplus \cdots \oplus  \tilde{\Psi}_n
    	\end{align}
    	Multiply by a linear mixing matrix.
    	\begin{align}
    	\Psi^{(\operatorname{out})}_{\tau_c} \leftarrow \tilde{\Psi}_{\tilde{\tau_c}} W^{\tilde{\tau_c}}{}_{\tau_c}
    	\end{align}
    \end{algorithmic}
    \end{algorithm}

    As our focus is on the construction of specific components in an SO(3)-equivariant architecture rather than on proposing an entirely new architecture,
    we demonstrate the potential of our formalism by incorporating it into existing neural networks.
    Specifically, we choose to augment the Cormorant architecture proposed in~\cite{Kondor_cormorant2019} and the recent state-of-art model~\cite{batatia2022designspace} with one additional three-body fusion block that replaces the conventional node-edge two-body interaction, with the aim of capturing inter-atomic interactions in a more faithful manner. Capturing three-body interactions in a SO(3) equivariant way with edge features could lead to a large overhead on computational resources.\risic{References missing}
    Applying fusion blocks to point clouds also requires ensuring that the resulting neural network obeys permutation symmetry. Since each fusion diagram has a single output, we can reinforce the permutation equivariance by passing these outputs through an aggregation function and incorporate them into existing message-passing-like mechanisms.
\textcolor{black}{ It is worth mentioning that except employing Clebsch-Gordan products, there are other efficient architectures like using spherical coordinates of neighboring atoms and leveraging spherical harmonics to encode angular momentum information into higher dimensional representation of SO$(3)$ and filtering through the spherical representations \cite{shuaibi2021rotation}. }

\subsection*{Cormorant with Fusion Diagrams (CoFD)}
    Cormorant is one of the first equivariant neural networks that utilize the group equivariance and designed to learn molecular dynamics simulations and ground-state molecular properties. \risic{This might be a bit of an exaggeration.} A neuron in Cormorant layer $s$ operates as follows:
    \begin{equation}\label{eq:vertex_activations}
        F^{s-1}_{i}= 
        \left[
        \underbrace{F^s_i \oplus \big(F^{s-1}_{i}\otimescg F^{s-1}_{i}\big)}_{\text{one-body part}} \oplus 
        \underbrace{ \Big(   \sum_{j} \Big(Y(\vec{x}_{ij}) \otimescg F^{s-1}_{j} \Big) g_{ij} \Big)}_{\text{two-body part}}\right]\cdot 
        W^{\text{vertex}}_{s,\ell}.
    \end{equation}
    Here $j$ sums over atom $i$'s local neighborhood, and
    $g_{ij}$ is a learned rotationally invariant function
    that takes $F^s_i$ and $F^{s}_j$ as input in addition to other rotationally invariant features. $\otimes_{\operatorname{cg}}$ denotes the channel-wise Clebsch-Gordan products and $Y(\vec{x}_{ij})$ the spherical harmonics with input the relative displacement vectors between $i$th and $j$th atoms
    (we refer the reader to~\cite{Kondor_cormorant2019} for the precise functional form of $g_{ij}$).
    Each of these terms corresponds to the two-body diagram on the left below,
    while the product with $g_{ij}$ is in a three-way product,
    it never has a covariant or contravariant component.

    In particular, we observe that this layer has no equivariant interaction between equivariant parts of the activation atom $i$ and the activation for atom $j$.
    Instead, their activations only interact through the rotationally invariant function $g$. Instead, in the present paper we employ our fusion diagrams to add an additional term to~\eqref{eq:vertex_activations}
    that fully integrates all of the information between atom $i$ and atom $j$.
    This corresponds to the fusion diagram on the right.
    The resulting fusion block has three inputs: the input activation for atom $i$, the input activation for atom $j$,
    and the collection of spherical harmonics evaluated on their relative displacements,
    and one output: a new feature for atom $i$.
    Consequently, we require a fusion diagram with three ingoing edges and one outgoing edge.
    Going from left to right, we input the representation of atom $i$, the representation of atom $j$, and the spherical harmonic representation of the edge connecting the two.
    We then incorporate this as a new term in~\eqref{eq:vertex_activations},
    giving the following functional form for our modified Cormorant layer. 
    \begin{equation}\label{eq:modded_cormorant}
        F^{s-1}_{i}= 
        \Big[
        F^s_i \oplus \big(F^{s-1}_{i}\otimescg F^{s-1}_{i}\big) \oplus 
        \sum_{i} \Big( \sum_{j} \Big( Y(\vec{x}_{ij}) \otimescg F^{s-1}_{j} \Big) g_{ij} \oplus F^{fusion}_{ij} \Big) \Big]\cdot 
        W^{\text{vertex}}_{s,\ell},
    \end{equation}
    where $F^{fusion}_{ij}$ is the output of the fusion block where inputs came from atom $i$ and atom $j$  within the coming legs chosen to be $i$th atom, $j$th atom, and their connecting edge. In other words, we use fusion diagrams to efficiently combine the atom-level messaging passing and edge-level message passing. 

\subsection*{MACE with Fusion Diagrams (MoFD)}

    \begin{figure}[H]
	\centering
	\includegraphics[width=1\textwidth]{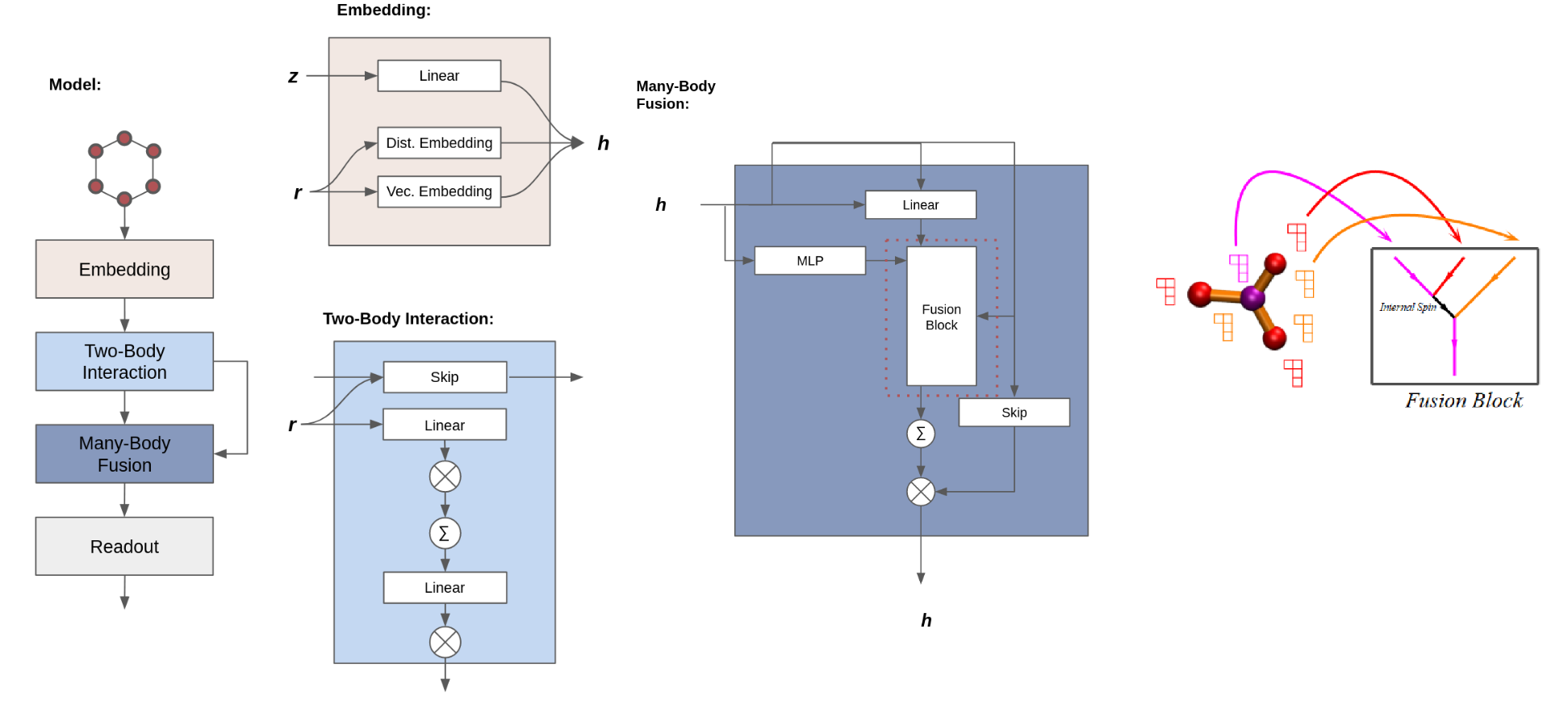}
	\caption{Schematic illustration of the implementation of fusion blocks in the MACE architecture. For each atom the fusion block first fuses all the neighboring atoms for a given radius cut-off by pre-selected fusion diagram templates. Specifically, for each neighboring atom, we fuse the information from the root, neighbor atom, and their connecting edge. Then the fusion block applies an aggregation method: in the present work, we simply sum all the neighbors.}
 
	\label{AtomsFusionBlocks}
\end{figure}
    \subsubsection*{Implementation of 3-body fusion blocks}\label{sparse}
    
    In our modified MACE architecture we use fusion diagrams (Figure \ref{AtomsFusionBlocks}), a local neighborhood is defined by a cut-off radius $r$, the information on the central particle is $\Psi^{(\operatorname{in, \vec{j}})}_{o}$, adjacent particle $\Psi^{(\operatorname{in, \vec{j}})}_{i}; i \in \mathcal{N}(o)$, and the incident edges $\Phi^{(\operatorname{in, \vec{j}})}_{(o, i)}$. In particular, this information is passed to a $l$-element sequence of linearly independent 3-body fusion interactions given by a sequence of different internal spin configurations $\{k_1, k_2, ..., k_l \}$, with the $l$-element sequence of outgoing activation on the center particle: 
    \begin{align}
        \Psi'^{(\operatorname{out}, \vec{j}_d)}_{o, k_{\alpha}} := \sum_{i \in \mathcal{N}(o)} \operatorname{FusionBlock} \left(\Psi^{(\operatorname{in, \vec{j}_a})}_{o} \Phi^{(\operatorname{in, \vec{j}_b})}_{i} \Psi^{(\operatorname{in, \vec{j}_c})}_{i} \right)
    \end{align}
    The final permutation invariant update to the center node information is obtained by concatenating $\{ \Psi'^{(\operatorname{out}, \vec{j})}_{o, k_{\alpha}} \}_{\alpha=1,..., l}$, followed by a linear mixing layer along the new concatenated axis. Note that in the sparse implementation, the feature dimension for each incoming activation never gets updated. Each time the internal spin configuration is only specified by a single internal spin label $k_\alpha$, thus sparsifying the three-body information flow. For each internal spin value $k$, the 3-body interaction fuses into a single SO($3$)-equivariant tensor (this fusion corresponds to the Fig.1 in the SM), while the final messaging passing aggregates neighboring edges and nodes information to the center node. In this implementation based on MACE architectures, we found our fusion block would only marginally increase the number of trainable parameters given the same channel width. 

    Fusion Block can also be initialized with significantly more trainable parameters than the original Mace does, which we denote as the \emph{dense} implementation. The key difference to the sparse implementation is the inclusion of multi-partite internal spins to create a nested $\operatorname{FusionBlock}$ module. More specifically, given a $l$-sequence of internal spins $\{k_{\alpha} \}_\alpha$, we can choose a tuple $(k_{\alpha_1}, k_{\alpha_2})$, a triple $(k_{\alpha_1}, k_{\alpha_2}, k_{\alpha_3})$, and beyond instead of specifying a single choice of the internal spin in the sparse implementation. Hence, a total of $l!$ selections can be made resulting in a significant boost to the model size and number of trainable parameters.  In our explicit implementation, we feed all choices of internal spins at once, resulting in a typical 10X boost of the trainable parameter size. As a result, we do not need to additionally pass a linear layer to reshape the channel width. The dense model could often outperform or be on par with the sparse implementation only with half the channel width. As an overall observation, the choices of internal spins are vital to our numerical performance. In our practice, the internal spins are chosen to range from $j = 0, 1, 2$, and sometimes with both parties.

\section*{Results} 
We describe three well-rounded benchmarks to test CoFD and MoFD, including QM-9\cite{ramakrishnan2014quantum} molecular property prediction, MD-17\cite{Chmiela2016a} small molecular dynamics, non-adiabatic molecular dynamics of stilbene. Our results are summarized in Figure \ref{fig:stilbene_scan}, Table \ref{tab:results} and Table \ref{tab:results_table}. 

    \subsection*{QM-9 Molecular properties and MD-17 molecular dynamics datasets}

    We first implement the fusion diagram on Cormorant architecture \cite{Kondor_cormorant2019}. 
    The standard QM-9 benchmark dataset\cite{ramakrishnan2014quantum} is used to test the performance of the CoFD model to predict molecular quantum properties of roughly 130,000 molecules in equilibrium, which contains multiple tasks of scalar value regression including atomization enthalpy, free energy, etc. In contrast, the MD-17 dataset \cite{Chmiela2016a} involves learning the ground-state PES and its gradient, for eight small organic molecules at room temperature from reference DFT calculations. 
    
    We compare the CoFD model and the original Cormorant model. The fusion diagram reduces the number of parameters in our networks, ensuring that we are not simply improving performance by adding additional parameters:
    for MD17, the networks with fusion diagrams have 135393 parameters compared to 154241 in the original Cormorant~\cite{Kondor_cormorant2019}, and our QM9 neural network has 121872 parameters compared to 299808 in the original~\cite{Kondor_cormorant2019}. \textcolor{black}{We report that the total time of training QM9 (resp. MD17) use 20 (resp. 12) hours with 256 Epoches, each with a mini-batch size of 64. Hence each epoch costs 281 (resp. 169) seconds.} Code for our modified network can be found at \url{https://github.com/ehthiede/diagram_corm}. \textcolor{black}{To be noted,   
the fusion block used in the CoFD to predict QM-9 and MD-17 is a sparse implementation. We did not use the dense implementation in predicting the QM-9 and MD-17 properties due to the large computational expense. However, it would be an interesting future direction to reduce the recourse overhead in the dense implementation, which would enable more subsequent experiments.   
}
    
    \begin{table}[t]
    \centering
    \caption{\label{tab:results} Mean absolute error of various prediction targets on QM-9 (left) and conformational energies (in units of kcal/mol) on MD-17 (right), for both the original Cormorant architecture and our modified version that incorporates a fusion block. It should be noted that the CoFD models have significantly fewer parameters than the original Cormorant. We report the mean and standard deviation from multiple runs. In comparison, the model with lower predictive error has been bolded. 
    }
    \begin{minipage}{0.49\textwidth}
    \small
    \setlength\tabcolsep{3pt}
\begin{tabular}{lrrrrr}
\toprule
{} &  Cormorant & &  CoFD &   &   \\
\midrule
$\alpha$ ($\mathrm{bohr}^3$) & 0.085&(0.001)& 0.088 & (0.003)&\\
$\Delta \epsilon$ (eV)       & 0.061&(0.005)& 0.062 & (0.001) &\\
$\epsilon_{\rm HOMO}$ (eV)   & \textbf{0.034}&\textbf{(0.002)}& 0.0391 & (0.0008) &\\
$\epsilon_{\rm LUMO}$ (eV)   & 0.038&(0.008)& 0.0347 & (0.0006)& \\
$\mu$ (D)                    & 0.038&(0.009)& 0.035 & (0.001)&\\
$C_v$ (cal/mol K)            & \textbf{0.026}&\textbf{(0.000)} & 0.0272 & (0.0002)& \\
$G$ (eV)                     &      0.020 &(0.000)& \textbf{0.0135} & (0.0002)& \\
$H$ (eV)                     &      0.021 &(0.001)& \textbf{0.0132} & (0.0004) &\\
$R^2$ ($\mathrm{bohr}^2$)    &      0.961 &(0.019)& \textbf{0.50} & (0.02) &\\
$U$ (eV)                     &      0.021 &(0.000)& \textbf{0.0130} & (0.0004)& \\
$U_0$ (eV)                   &      0.022 &(0.003)& \textbf{0.0133} & (0.0003) &\\
ZPVE (meV)                   &      2.027 &(0.042)& \textbf{1.43} & (0.04) & \\
\bottomrule
\end{tabular}

    \end{minipage}
    \hspace{1 mm}
    \begin{minipage}{0.49\textwidth}
    \small
    \setlength\tabcolsep{3pt}
\begin{tabular}{lrrrrrr}
\toprule
{} &  Cormorant &  CoFD   \\
\midrule
Aspirin        &\     0.098 &   \textbf{0.0951}  \\
Ethanol        &     0.027 &   \textbf{0.0241}  \\
Malonaldehyde  &     0.041 &   \textbf{0.0380}  \\
Naphthalene    &\textbf{      0.029 }&   0.0321  \\
Salicylic Acid &      0.066 &   \textbf{0.0608 } \\
Toluene        &      0.034 &   \textbf{0.0316 } \\
Uracil         &\textbf{      0.023 }&   0.0297  \\
\bottomrule
\end{tabular}
    \end{minipage}
    \end{table}

\subsection*{Stilbene Non-adiabatic molecular dynamics}

Non-adiabatic MD (NAMD)\cite{tully2012perspective} is a powerful approach for predicting photo-induced chemical processes, including photo-catalytic reactivity\cite{tong2012nano}, photo-induced DNA damage\cite{dougherty1998photodynamic}, and the performance of sun-screening products\cite{baker2017photoprotection}. Unlike ground-state dynamics, NAMD involves evaluating multiple PESs and their gradients simutaneously. However, studying excited-state dynamics requires higher accuracy electronic structure methods than DFT\cite{mai2019influence}, resulting in significantly higher computational costs. Thus, there is motivation to test our model's ability to study multiple PESs that are not generated by DFT.

\begin{table}[ht]
\centering
\resizebox{1\columnwidth}{!}{
\begin{tabular}{@{} lllllll @{}}
\toprule
Model & Feature Dimension & Num. of Param. & Train Size & Ground State (Energy, Forces) & First Excited State (Energy, Forces) & Second Excited State (Energy, Forces) \\
\midrule
MACE & 64 & 330320 & 285 & (\textbf{19.15}, 0.70) & (\textbf{9.88}, 1.25) & (\textbf{21.80}, \textbf{1.03}) \\
MoFD-sparse & 16 & 66784 & 285 & (19.74, \textbf{0.62}) & (13.43, \textbf{1.12}) & (23.42, 1.07) \\
MACE & 128 & 979088 & 950 & (\textbf{26.44}, 1.30) & (\textbf{29.07}, 3.56) & (\textbf{48.77}, 3.05) \\
MoFD-sparse & 32 & 141168 & 950 & (28.84, 1.40) & (36.69, 3.55) & (55.08, 3.11) \\
MoFD-dense & 16 & 690976 & 950 & (27.59, \textbf{1.14}) & (32.64, \textbf{3.31}) & (54.76, \textbf{2.65}) \\
\bottomrule
\end{tabular}
}
\caption{Comparative analysis of MACE and MoFD models in dense and sparse implementations, evaluated on single and multiple independent non-adiabatic trajectories of cis-stilbene. The table presents the feature dimension, number of parameters (Num. of Param.), training set size (Train Size), and results for the ground state, first excited state, and second excited state. Results include energy values in milli-Hartree (mHartree) and forces in milli-Hartree per Angstrom (mHartree/A). The bold figures represent the best performance in each category.}
\label{tab:results_table}

\end{table}

\begin{figure}[h]
\centering
\includegraphics[width=1\textwidth]{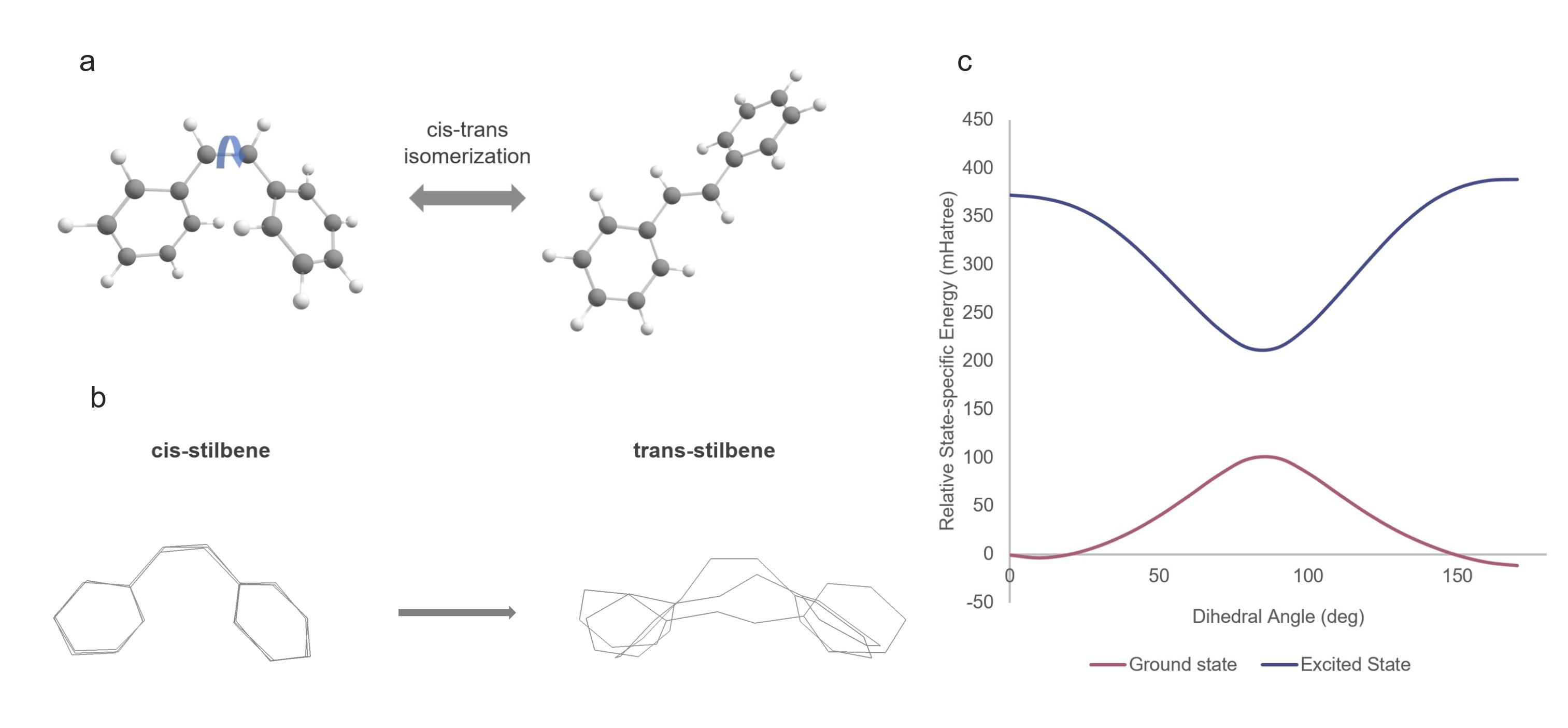}
\caption{(a) Illustration of photo-induced cis-trans isomerization of stilbene (b) Initial and end configurations of three representative trajectories, which are Wigner-sampled.\cite{dahl1988morse} (c) the one-dimensional cut of stilbene ground/ excited-state PESs by rotating the carbon-carbon bond as illustrated in (a), which illustrates the energetic diagram of stilbene isomerization process.}

\label{fig:stilbene_scan}
\end{figure}

In this study, we explore the photo-induced cis-trans isomerization process of stilbene, a phenomenon first reported by Syage \cite{syage1982picosecond}. Our approach utilizes the Complete Active Space Self-Consistent Field (CASSCF) theory \cite{roos1980complete}, specifically targeting the conjugated $\pi$ orbital localized on the carbon-carbon double bond and its anti-bonding counterpart. This selection forms our active space, characterized as two electrons in two orbitals (2e,2o), and all calculations are conducted using the 6-31G* basis set. To accurately capture the quantum effects inherent in photoisomerization, we adopt a quantum-classical approximation through trajectory surface hopping (TSH), as implemented in the SHARC package \cite{mai2018nonadiabatic}. This method integrates both quantum and classical dynamics, crucial for studying processes like isomerization. Wigner sampling \cite{BiedenharnUnWigner1967} is employed to generate a variety of initial configurations, initiating the molecular trajectories under study.

A stringent criterion is applied to ensure the quality of the data: only trajectories maintaining total energy conservation within 0.2 eV were considered valid and included in the dataset. This threshold ensures the physical relevance of the trajectories by excluding those that do not adhere to energy conservation principles. The resultant dataset, therefore, comprises multiple molecular trajectories of stilbene, predominantly initiated in an excited state. These trajectories provide a comprehensive view of the isomerization process, offering valuable insights into the dynamics of this photochemical reaction. Detailed computational specifications and a more thorough introduction to the methods employed are available in the SM.

The widely-adopted MD17 dataset \cite{chmiela2017machine} comprises adiabatic dynamic trajectories using the PBE functional, though the spin polarization, basis set, and computational grid information are absent from the literature, near equilibrium, where molecular movements are trivial. As a result, MD17 is heavily biased towards sampling the reactant region of the PES without considering the driven non-equilibrium forces.\cite{pengmei2023xxmd} However, a meaningful chemical reaction typically involves three parts on the PES: reactant, product, and transition state. It is important to note that the accuracy of common density functionals is usually a few kcal/mol when compared to higher levels of theory. For example, the PBE functional used in the MD17 dataset has an average error of more than 9 kcal/mol (roughly 0.4 eV) when predicting reaction barriers \cite{mardirossian2017thirty}. In contrast, the trajectories we sampled visited the reactant, product, and transition state regions of multiple PESs, as illustrated in Figure \ref{fig:stilbene_scan}.

To compare the performance of MACE and MoFD with sparse implementation, we selected one reactive trajectory and employed the MACE model with a feature channel dimension of 64 and high-order equivariant features with $l=0,1,2,3$. For MoFD, we maintain the same feature angular momentum and set the feature channel dimension to 16, resulting in a model with only 66,784 parameters, an order of magnitude smaller than that of MACE. Given the increased difficulty in predicting atomic forces, we adjust the training loss on energy and forces with a ratio of 1:1000, as recommended in previous literature \cite{batzner2021e3equivariant,batatia2022mace}. As the loss is disproportionately weighted towards the force, we concentrate on the force regression performance. The models is trained on 285 samples and tested on a separate hold-out test set of 428 samples. The models were trained in a state-specific fashion, which means each model regress single state's PES and forces for comparison purposes. \textcolor{black}{ Our findings indicate that MoFD with sparse implementation has a decent performance in force prediction for the first two states, while MACE fits better when predicting the energy.}

We further assess the generalization ability of our models across different trajectories by incorporating two additional independent trajectories into the dataset, resulting in a total of 950 training samples and 1,395 hold-out testing samples. We increase the complexity of MACE by expanding its feature channel width to 128, leading to a total of 979,088 parameters. Concurrently, we double the feature dimension of MoFD to 32, making it only as large as MACE. Additionally, we implement MoFD with a dense feature dimension of 16, with equivariant features $l=0,1,2$, resulting in a total of 690,976 parameters (29.4\% fewer than the original MACE). \textcolor{black}{In terms of runtime, each epoch requires 52 seconds in the MACE model compared to 31 seconds in the MoFD model, attributed to the utilization of lower-dimensional angular momentum features as inputs.} The MoFD model with the dense implementation surpass MACE in the force prediction tasks, while the MoFD model with the sparse implementation remains comparable to MACE's accuracy as indicated in Table \ref{tab:results_table}.
Nonetheless, it is crucial to note that the performance of all models decrease when learning excited states due to the less well-defined topologies of excited-state PESs \cite{mai2019influence}.

\section*{Discussion}

In this work, we have introduced a new method for constructing equivariant blocks for rotation-equivariant layers based on fusion diagrams.
Previous work has shown that tensor products can be used to construct neurons for rotation-equivariant neural networks. Moreover, prior research has observed that neural network ansatzes for the quantum system can be unified with spin network ansatzes. Our work is the first to employ these connections in the opposite direction: by employing diagrammatic methods used in physics, we construct new components that can be incorporated into equivariant neural networks. 

Using classic results from invariant theory, we show that neural networks built from using fusion blocks are capable of approximating any continuous SU(2)-equivariant functions. To demonstrate the practical utility of fusion blocks, we perturb existing SO(3) equivariant neural network architectures, such as Cormorant\cite{Kondor_cormorant2019} and MACE\cite{batatia2022mace}, by incorporating a fusion block in each layer. The modified architectures generally achieves better performance for a smaller number of parameters. Indeed, the idea of using equivariance and symmetry to prune neural networks has been applied \cite{wang2022symmetric} in the quantum setting. We believe this indicates that fusion blocks can be a useful addition to group-equivariant neural networks. 

To test the performance of the fusion block approach, we apply the revised CoFD and MoFD models not only to the standard benchmark datasets QM-9\cite{ramakrishnan2014quantum} and MD-17\cite{Chmiela2016a}, but also novel applications such as non-adiabatic molecular dynamics. We find that the addition of the fusion blocks improved the performance of the models. 


In future work, we hope to use fusion blocks to improve the interpretability of equivariant neural networks. In theoretical physics, fusion diagrams represent physical processes that correspond to many-body interactions. Furthermore, physicists often manipulate fusion diagrams through internal permutations through a process known as \emph{recoupling}. Recouplings relate to the physical properties of different fusion diagrams and can show symmetries present in the products that may not be immediately apparent by inspection. 
Employing the formalism of recoupling may highlight hidden symmetries in the network architecture, indicating new ways to save computational effort. Employing the language of fusion diagrams in these settings could help unify our physical picture of fusion diagrams with computational realities. Finally, fusion diagrams are graphical representations of ways in which local atoms are being fused. It is of interest to consider the effect of the local subgraph topology on the corresponding fusion blocks; in particular, whether fusion diagrams serve as a general principle towards building more expressive graph neural nets with 3D equivariance specific to chemical applications. We leave addressing these questions as future research opportunities.

\bibliography{sample}

\section*{Acknowledgements }

J.L. is supported in part by International Business Machines (IBM) Quantum through the Chicago Quantum Exchange, and the Pritzker School of Molecular Engineering at the University of Chicago through AFOSR MURI (FA9550-21-1-0209).



\section*{Supplementary material}

# Supplementary Material

Zimu Li, Zihan Pengmei, Han Zheng, Erik Thiede, Junyu Liu, Risi Kondor

# Contents

# 1 Overview of Fusion Diagrams

Clebsch-Gordan products (CG-products) appear naturally in quantum mechanics out of necessity to describe the interaction of multiple particles under global SU(2) symmetry. For even moderately complex quantum mechanical systems, equations keeping track of successive products will become formidably complicated. Consequently, physicists have developed a diagrammatic representation of successive Clebsch-Gordan decompositions. We refer to these diagrams as *fusion diagrams*. The space of fusion diagrams that fuse $n$ incoming SU(2)-contravariant features $\{\psi^{(j_i)}\}_{i=1}^n$ (defined in Background in the main text) into a single outgoing SU(2)-contravariant feature $\Psi^{(J)}$ comprised of successive Clebsch-Gordan products is denoted by $\mathrm{Hom}_{\mathrm{SU}(2)}(V_{j_1} \otimes \cdots \otimes V_{j_n}, V_J)$ (Fig.1). By choosing a fusion diagram $Q^{(j_1,\ldots,j_n;J)} \in$

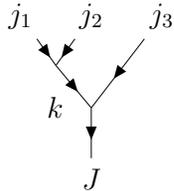

Figure 1: A fusion diagram as a standard basis element in $\mathrm{Hom}_{\mathrm{SU}(2)}(V_{j_1} \otimes V_{j_2} \otimes V_{j_3}, V_J)$. Fusion diagrams of this type with all possible values $k$ are also called Wigner 4-$jm$ elements.

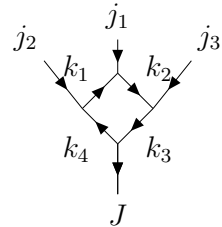

Figure 2: Another fusion diagram as some element in $\mathrm{Hom}_{\mathrm{SU}(2)}(V_{j_1} \otimes V_{j_2} \otimes V_{j_3}, V_J)$.

$\text{Hom}_{\text{SU}(2)}(V_{j_1} \otimes \cdots \otimes V_{j_n}, V_J)$ the fusion process can be computed by the following form like inner products:

$$\psi^{(J)} = \langle Q^{(j_1, \cdots, j_n; J)}, \psi^{(j_1)} \otimes \cdots \otimes \psi^{(j_n)} \rangle. \tag{1}$$

To be more specific, in a fusion diagram, e.g., Fig.1 & 2, each edge is associated with a spin label, which is a non-negative integer or half-integer standing for an SU(2) irrep. While only integers appears in the case of SO(3) irreps. Each vertex represents a common CG-product $C^{(j_a, j_b, j_c)}$ and is therefore connected to exactly three edges [1]. We adopt the following convention in labeling fusion diagrams, which have $n$ external incoming edges and one external outgoing edge:

1. Each fusion diagram has a list of external spin labels $\{j_1, j_2, ..., j_n\} \equiv \mathbf{j}$, and a list of internal spin labels $\{k_1, ..., k_m\} \equiv \mathbf{k}$. These lists are referred to as external and internal spin configurations respectively.
2. For each vertex, input vectors are always assumed to be transformed contravariantly, i.e., with upper entry indices $m$ so that they are fused together via Eq.(1) and yield a contravariant output which would be used for subsequent updates of CG-products like in Figure 1.
3. The spins associated with each vertex must obey the Clebsch-Gordan rule: denoting the two incoming edges $j_a$ and $j_b$ and the outgoing edge $j_c$, we require that $|j_a - j_b| \le j_c \le j_a + j_c$ [3, 25].
4. We label the final outgoing edge with $J$ so that incoming $n$ SU(2)-contravariant features are fused into one SU(2)-contravariant feature to be processed in the next layer.

For instance, the fusion diagram in Figure 1 corresponds to

$$Q^{(\mathbf{j}, \mathbf{k}; J)}{}_{m_{j_1} m_{j_2} m_{j_3}}{}^{m_J} \equiv M\, C^{(j_1 j_2 k)}{}_{m_{k_1} m_{j_2}}{}^{m_k} C^{(k j_3 J)}{}_{m_k m_{j_3}}{}^{m_J}, \tag{2}$$

where $M$ is a normalization coefficient equal to $\sqrt{(2k_1 + 1)(2k_2 + 1)}$.

In the following discussion, we will refer to any fusion diagrams that form a basis in the invariant Hom set as *Wigner n-jm symbol* where $n$ counts the total outgoing and incoming edges. Wigner $n$-$jm$ elements such as Fig.1 are prototypes for fusion blocks and essential in proving our theoretical results later. For a fixed set of external spin labels $\mathbf{j}$, varying all possible assignments of $\mathbf{k}$ of (2) exhausts all possible orthogonal basis elements in $\text{Hom}_{\text{SU}(2)}(V_1 \otimes V_2 \otimes \cdots \otimes V_n, U)$ [25, 16, 4].

There are also more general spin diagrams that we can draw. For instance, with three incoming external edges and one outgoing edge (Figure 2)

$$Q^{(\mathbf{j}, \mathbf{k}; J)}{}_{m_{j_1} m_{j_2} m_{j_3}}{}^{m_J} = C^{(j_1, k_2, k_1)}{}_{m_{j_1}}{}^{m_{k_2}}{}_{m_{k_1}} C^{(j_2 k_1 k_4)}{}_{m_{j_2}}{}^{m_{k_1}}{}_{m_{k_4}} C^{(j_3, k_3, k_2)}{}_{m_{j_3}}{}^{m_{k_3}}{}_{m_{k_2}} \\ C^{(J, k_4, k_3) m_J m_{k_4}}{}_{m_{k_3}}, \tag{3}$$

can be expressed as a linear combination of Wigner-4jm tensors in (2) with different internal spin configuration $\mathbf{k}$ because they form a basis for all such high-order CG-products. [2] Note that the dimension of the invariant Hom set grows exponentially with the number of outgoing and incoming edges so it remains an intriguing question to determine which fusion diagrams could span larger subspaces that help build practically useful universal neural network architectures. One motivation for using more general fusion diagrams such as Fig.2 is that not only does it spans multiple Wigner symbols at once, it can encode non-trivial local graphical invariance

---

[1] One can also assume that vertices connected with more than three edges are sup-fusion diagrams with vertices having only three edges. We refer the interested audience to[23, 22, 21] for more details. Note for general tensor networks without prescribed symmetry, there is no such restriction.

[2] For readers familiar with matrix product states, we observe that (3) is a matrix product state with boundary conditions used in tensor networks [23, 22, 21].

such as the $C_4$ group invariance when choosing appropriate **j** and **k**.

## 2   Proof of the Universality Theorem

### Equivariant Universal Approximation

Rotationally equivariant neural architectures affording universal approximation on point clouds are analyzed in [15]. Here we show that the fusion block activation is an universal approximator subject to SU(2) or SO(3) equivariance but using different methods, borrowing techniques from classical invariant theory, which studies invariant and equivariant polynomials of classical groups [18, 20, 8]. Similar universality statements to the ones we present have also been given in [4]. We pursue a different proof strategy, using iterated Clebsch-Gordan products instead of constructing all basis elements for the infinite-dimensional space of equivariant functions. To be specific, it was proved by Gordan, Hilbert, and others in Classical Invariant Theory that Clebsch-Gordan products are *the only* essential ingredients to construct any SU(2)-equivariant function [10, 18, 20]. A formal statement of this significant result is provided in Theorem 2.2 later while we present the following adapted form for our case.

**Theorem 2.1.** *Let $V_j, W_{j'}$ be two irreps of SU(2) and let $\Psi_1, ..., \Psi_n \in V_j$ be spin-j vectors as inputs. For simplify, we temporarily denote by $(\Psi_r, ..., \Psi_s)$ an iterated CG-product. Given a multi-degree $\mathbf{d} = (d_1, ..., d_n)$, a CG-product for multi-homogeneous monomial of degree $\mathbf{d}$ is written as*

$$\left( \underbrace{(\Psi_1, ..., \Psi_1)}_{d_1 \text{ copies}}, \cdots, \underbrace{(\Psi_n, ..., \Psi_n)}_{d_n \text{ copies}} \right) \tag{4}$$

*Any SU(2)-equivariant polynomial $f: V_j^{\oplus n} \to W_{j'}$ can be written as a linear combination of these multi-homogeneous monomials.*

The above theorem says that any SU(2)-equivariant polynomial can be constructed from CG-products. Using an equivariant version of the Stone-Weierstrass theorem [5], that any continuous equivariant function can be approximated by equivariant polynomials. As a result, they can be approximated by iterated CG-products which are graphically represented as fusion diagrams in our setting.

We now explore both rotation and permutation equivariant universality of neural networks using fusion blocks. Graph neural networks (GNNs) in practice perform local aggregation on the graph; in our case, this corresponds to the case where the fusion blocks may be applied only to particles in a local atomic neighborhood. For this reason, we prove universality for functions computed in the vicinity of an atom in the points cloud. This is a standard and reasonable assumption since the the graph structure of points clouds are defined based on interactions of atoms within a radial distance for which the induced local subgraph is fully connected. If receptive field is set to be global, we obtain the universality when taking interactions of any pair of atoms from the points cloud into account. Let us assume that the learning task is defined on $V_j^{\oplus n}$ with $n$ representing the number of atoms in a local neighborhood or simply the entire system. Intuitively, as fusion blocks fuse inputs on each atom with the center and then aggregate (see Fig.1 in the main text), iterated CG-products will be built with permutation equivariance after updating several layers.

The universality theorem is proved by results from Classical Invariant Theory [9, 18, 20] and Quantum Angular Momentum Recoupling Theory [14, 25, 16]. Classical Invariant Theory studies polynomial maps from $V$ to $W$ equivariant under classical group actions, like these of SO(n), SU(n), SL(n,$\mathbb{C}$). For short, they are called *G-equivariant polynomials* or *covariants*.

For instance, linear equivariant maps commonly used in designing equivariant neural networks are simply equivariant polynomials of degree one. Clebsch-Gordan products like $C^{j_1j_2k}\psi^{j_1}\psi^{j_2}$ are equivariant polynomials of degree two. Invariant polynomials are simply special cases of covariants when the target space $W$ admits the trivial $G$-representation. Quantum Angular Momentum Recoupling Theory specializes the case for SU(2) tensor product representations. Different ways of taking tensor products are referred as different *recoupling schemes* and produces different results in computing Clebsch-Gordan coefficients (CGC) which has real significance in studying quantum system like the $jj-LS$ recoupling for electron configuration [14]. Spin diagrams used in this paper are graphical interpretation of various recoupling schemes. They are further generalized in Topological Quantum Computation [11] and Quantum Gravity [19]. Classical Invariant Theory and Angular Momentum Recoupling Theory provides different insights on Clebsch-Gordan products. We will introduce gradually how we benefit from these insights and obtain the universality theorem in the following.

## 2.1 Equivariant Polynomials and Clebsch-Gordan Product

Equivariant layers are central part in equivariant neural networks. Formally, they are functions $L$ that commutes with the group action

$$T'_g \circ L = L \circ T_g, \forall g \in G.$$

Most previous researches only focus on *linear equivariant layer* which is a linear map commuting with the $G$-action (e.g., a convolution). However, one can in principle design *nonlinear equivariant layer* from the very beginning by equivariant polynomials. As a standard notation from Classical Invariant Theory, we denote by $P(V,W)^G$ the collection of all equivariant polynomials from $V$ to $W$ and $P(V)^G$ the collection of all invariant polynomials on $V$.

To determine SU(2)-equivariant functions from $V$ and $W$, we decompose $V, W$ into irreps and consider the problem within each irrep. As a well-known result from representation theory, *Schur's lemma* states that linear equivariant maps between isomorphic irreps are simply scalars. For non-isomorphic irreps, the only possible linear equivariant map is the trivial zero map [8]. Therefore, in our setting with $V = V_j^{\oplus n}$, linear equivariant layers can only be defined as learnable weights combining $n$ inputs $(\psi_1, ..., \psi_n)$ linearly [12]. To update inputs in a nonlinear manner, we can employ the *Clebsch-Gordan decomposition* or *Clebsch-Gordan product* which is formally defined from the tensor product $V_{j_1} \otimes V_{j_2}$ of two irreps to a third irrep $V_J$ which can be decomposed from the tensor product:

$$C^J_{j_1j_2} : V_{j_1} \otimes V_{j_2} \to V_J.$$

The Clebsch-Gordan non-linearity was first used in [13, 1]. Explicitly, for any $(\psi_1, \psi_2) \in V_{j_1} \oplus V_{j_2}$ expanded under basis (spherical harmonics basis/quantum angular momentum basis),

$$\left(C^J_{j_1j_2}(\psi_1, \psi_2)\right)^M = (C^J_{j_1j_2})^M{}_{m_1m_2} \psi_1^{m_1} \psi_2^{m_2},$$

where $m_i$ denotes the magnetic number, is a *homogeneous polynomial of degree two* because we have the quadratic term $\psi_1^{m_1}\psi_2^{m_2}$ of inputs.

On the other hand, one may argue that the CG-product is linear on each input. That is correct and a CG-product should be formally called a *bilinear function*. We denote the collection of SU(2) bilinear functions by $\text{Hom}_{\text{SU}(2)}(V_{j_1}, V_{j_2}; W)$. Another equivalent notation $\text{Hom}_{\text{SU}(2)}(V_{j_1} \otimes V_{j_2}; W)$ and its generalization $\text{Hom}_{\text{SU}(2)}(V_{j_1} \otimes \cdots \otimes V_{j_n}; W)$ for multilinear functions are used more often in the main article. Mathematically, bilinear map on the direct sum $V_{j_1} \oplus V_{j_2}$ corresponds uniquely to a linear map on the tensor product $V_{j_1} \otimes V_{j_2}$ and hence we

use the above notation. This is also the reason that even originally defined on tensor products, we are free to use CG-products on direct sums without any ambiguity. As a reminder, assume $j_1 = j_2 = J$, the following map

$$V_j \mapsto V_j \oplus V_j \to V_J; \quad \psi \mapsto (\psi, \psi) \mapsto C^J{}_{jj}(\psi, \psi).$$

is non-linear and equivariant from $V_j$ to $V_J$. It circumvents the restriction of Schur lemma on linear equivariant maps.

To discuss more nontrivial examples, we now present two ways to generalize CG-products to high order as the central notions used for the proof:

1. *Iterated CG-series* is a canonical high order CG-product by contracting common CG-products sequentially, e.g.,

$$C^J{}_{kj_3} C^k{}_{j_1j_2} : V_{j_1} \oplus V_{j_2} \oplus V_{j_3} \to V_k \oplus V_{j_3} \to V_J;$$
$$(\psi_1, \psi_2, \psi_3) \mapsto \left(C^k{}_{j_1j_2}(\psi_1, \psi_2), \psi_3\right) \mapsto C^J{}_{kj_3}\left(C^k{}_{j_1j_2}(\psi_1, \psi_2), \psi_3\right)$$

An *nth iterated CG-product* is an equivariant polynomial of degree $n+1$. It is also called a *multi-linear function* with $n$ inputs living in $\mathrm{Hom}_{\mathrm{SL}(2,\mathbb{C})}(V_{j_1} \otimes \cdots \otimes V_{j_n}; W)$.

2. Given any iterated CG-product $Q$, suppose $(j_1, ..., j_k)$ are all distinct spin labels of $Q$ but each of which appears $d_i$ times (like $C^J{}_{jj}$ from above). Inputting the same vector for all repeated spin labels defines a *multi-homogeneous polynomial* of degree $d = (d_1, ..., d_k)$. For instance

$$C^J{}_{j_1k}\left(C^k{}_{j_2j_1}(\psi_1, \psi_2), \psi_1\right)$$

is a multi-homogeneous polynomial of degree $(2, 1)$. One variable polynomial can be decomposed into homogeneous polynomials, as a generalization, multi-homogeneous polynomials are building blocks for multivariate polynomials [20]. At a result, one cannot always expect a polynomial to be linear on each of its variable.

The above examples shows some hints that CG-products can be assembled together to construct any complicated SU(2)-equivariant polynomial. This is true due to Gordan Theorem [9, 18, 17]:

**Theorem 2.2** (Gordan Theorem). *Given a complex SL(2, $\mathbb{C}$) representation $V = \bigoplus V_j^{\oplus m_j}$ ($m_j$ denotes possible multiplicity of $V_j$), any equivariant polynomial $f$ on $V$ can be written as a linear combination of iterated CG-products $((...((\psi_1, \psi_2)_{k_1}, \psi_3)_{k_2}, ..., \psi_n)_{k_{n-1}}$. Note that each input may occur more than once in the iteration.*

We use the simplified notation for Iterated CG-products in the above theorem. That is the most standard of notation used in classical invariant theory where CG-product is called *transvectant*. Transvectants are expressed by polynomials and differential operators by Gordan, Hilbert and others [10, 18], while the matrix/tensor notation of CG-products was adapted by later researchers like Wigner and Racah to study quantum physics and representation theory [14, 25]. We follow the tensor notation from physics and representation theory in this paper.

Gordan theorem was originally formulated to find all *generators* of SL(2, $\mathbb{C}$)-equivariant polynomials, which can be further restricted to SU(2)-equivariant polynomials by definition. It was then proved by Hilbert in his celebrated Basis Theorem that one just needs finitely many generators to generate the infinite dimensional collection of equivariant polynomials (see [18, 8] for more details). Even though it is extremely difficult to write down these generators explicitly. For small spin labels ($j \sim 10$), a few explicit generators of SL(2,$\mathbb{C}$) irreps are summarized in [18, 17]. The so-called First Fundamental Theorem says that generators of

fundamental representation of GL(n), SO(n) and Sp(n) are, loosely speaking, inner products [8] (inner products are fundamental cases of CG-products). For this reason, we do not bother ourselves to compute generators but simply use Gordan theorem which is enough to claim that CG-products are *the only* essential ingredient to construct any SU(2)-equivariant polynomial.

## 2.2 Spin Diagram and Spin Recoupling

Except the above two algebraic expressions of CG-products: tensor notation and transvectant, we present more details about *spin diagrams/fusion diagrams* as the main topic of this paper. In the context of physics, iterated CG-products are special cases of tensor networks with SU(2)-symmetry [23, 22, 21]. While general tensor networks are obtained by contracting arbitrary tensors which may not be SU(2)-equivariant [3]. Graphically, we draw

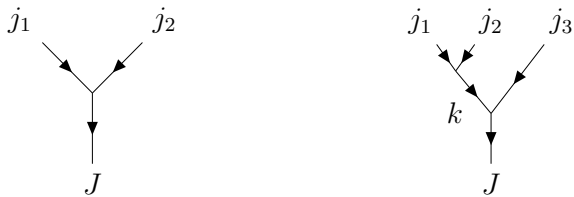

as a common CG-product (LHS) and a third order iterated CG-products (RHS) with $k$ being an internal spin. Outputs of fusion diagrams are generally spin-$J$ vectors. To construct *invariant polynomials* with scalar output however, one can set $J=0$ or consider spin diagrams like

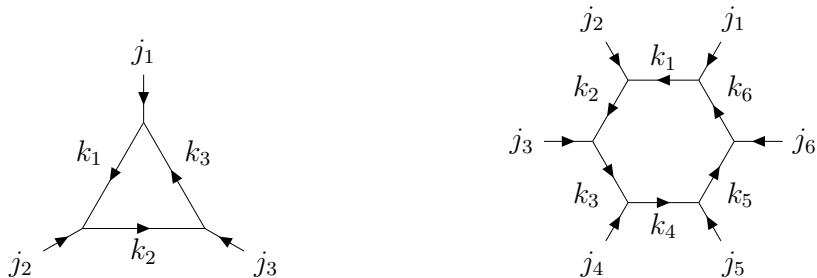

To build even complicated fusion diagrams, we just need to keep in mind that each vertex in the diagram is connected with *three and only three legs* which represents a simple CG-product (see [23, 22, 21] for more examples). Based on the graphical expression, Gordan theorem 2.2 says that no matter how complicated a fusion diagram would be, it can be written as a linear combination of fusion diagrams of iterated CG-products.

As we mentioned above, multi-homogeneous polynomials are building blocks for multivariate polynomials. Because of repeated inputs, a general multi-homogeneous polynomial of degree $d = (d_1, ..., d_k)$ has $\frac{d!}{d_1!\cdots d_k!}$ possible different forms even written as a iterated CG-product like

$$C^J_{j_2 k}\left(C^k_{j_1 j_1}(\psi_1, \psi_1), \psi_2\right), \quad C^J_{j_1 k}\left(C^k_{j_2 j_1}(\psi_2, \psi_1), \psi_1\right), \quad C^J_{j_1 k}\left(C^k_{j_1 j_2}(\psi_1, \psi_2), \psi_1\right).$$

The notion of *spin recoupling* accompanied spin diagrams from angular momentum recoupling theory can help to resolve this problem: like an extension of Gordan theorem, it says that for instance, the second and third iterated CG-products from above can be written as a linear combination of the first one with internal spins varying. This result is desirable because it facilitates the proof of universality theorem later. To be precise, recoupling can be summarized

---
[3]It is more common to use the word "symmetry" to indicate that a physics system is symmetric under certain transformation while the word "equivariance" is more formal in group theory

as two symmetric properties when computing Clebsch-Gordan coefficients [14, 25, 11]:

1. The *R-move*:

$$(C^J_{j_1j_2})^M_{m_1m_2} = (-1)^{j_1+j_2-J}(C^J_{j_2j_1})^M_{m_2m_1}$$

which says if we exchange the coupling order from $V_{j_1} \otimes V_{j_2}$ to $V_{j_2} \otimes V_{j_1}$, Clebsch-Gordan coefficients will differ by a phase factor $(-1)^{j_1+j_2-J}$. It is symmetric in some cases (like inner products), but alternating in other cases. Graphically, we draw

$$\begin{array}{c}j_1 \quad\quad j_2 \\ \searrow\swarrow \\ \downarrow \\ J\end{array} \quad = \quad (-1)^{j_1+j_2-J} \quad \begin{array}{c}j_2 \quad\quad j_1 \\ \searrow\swarrow \\ \downarrow \\ J\end{array}$$

2. The so-called *F-move* accounts for higher order cases: consider the following third order iterated CG-products:

$$(\psi_1, \psi_2, \psi_3) \mapsto \bigl(C^k_{j_1j_2}(\psi_1, \psi_2), \psi_3\bigr) \mapsto C^J_{kj_3}\bigl(C^k_{j_1j_2}(\psi_1, \psi_2), \psi_3\bigr),$$
$$(\psi_1, \psi_2, \psi_3) \mapsto \bigl(\psi_1, C^l_{j_2j_3}(\psi_2, \psi_3)\bigr) \mapsto C^J_{j_1l}\bigl(\psi_1, C^l_{j_2j_3}(\psi_2, \psi_3)\bigr),$$

where the first operator couples $j_1, j_2$ firstly and then takes $j_3$ while the second couples $j_2, j_3$ firstly and then takes $j_1$. These two coupling schemes can be transformed by *Winger 6j-symbols*:

$$(C^J_{j_1l}C^l_{j_2j_3})_{m_1m_2m_3M} = \sum_k (2k+1)(-1)^{j_2+j_3+k+l}\begin{Bmatrix} j_1 & j_2 & k \\ J & j_3 & l \end{Bmatrix}(C^J_{kj_3}C^k_{j_1j_2})_{m_1m_2m_3M}.$$

It can be though as a generalized associative rule when coupling more than two spins. Graphically, we draw

$$\begin{array}{c}j_1 \quad j_2 \quad j_3 \\ \searrow\swarrow \swarrow \\ k \\ \downarrow \\ J\end{array} \quad = \quad \sum_l \begin{Bmatrix} j_1 & j_2 & k \\ J & j_3 & l \end{Bmatrix} \quad \begin{array}{c}j_1 \quad j_2 \quad j_3 \\ \searrow\searrow\swarrow \\ l \\ \downarrow \\ J\end{array}$$

Back to the above question on multi-homogeneous polynomials, we see that

$$C^J_{j_1k}\bigl(C^k_{j_1j_2}(\psi_1, \psi_2), \psi_1\bigr) = \sum_k (2k+1)(-1)^{j_2+j_3+k+l}\begin{Bmatrix} j_1 & j_2 & k \\ J & j_1 & l \end{Bmatrix} C^J_{j_2l}\bigl(C^l_{j_1j_1}(\psi_1, \psi_1), \psi_2\bigr).$$

Converting the third iterated CG-product to the first from above is more tricky, we illustrate by the following manipulation of fusion diagrams with R-move and F-move (summations are

omitted):

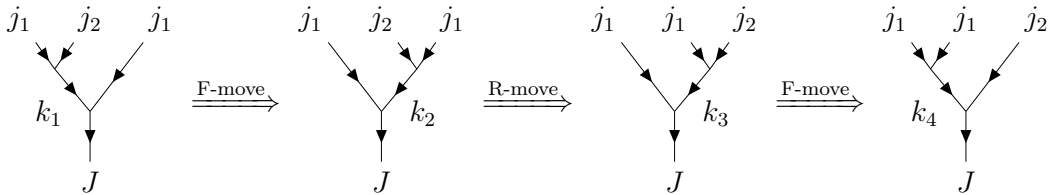

There is no need to compute 6j-symbols. Our aim is just verifying theoretically that any complicated multi-homogeneous polynomial written by iterated CG-products can be standardized as above. A precise standardization scheme is given in the next subsection (Definition 2.3).

## 2.3 Equivariant Universality with Fusion Blocks

We now prove the universality of fusion blocks. Let us first make a formal definition:

**Definition 2.3.** Let $V_j, W_{j'}$ be two irreps of SU(2) and let $\psi_1, ..., \psi_n$ be spin-$j$ vectors as inputs. Given a multi-degree $d = (d_1, ..., d_n)$, we take fusion diagrams coupling each $\psi_i$ by $d_i$ times. Then we map the processed $n$ vectors into a iterated CG-product. The whole fusion diagram is called a *standardized Clebsch-Gordan product for multi-homogeneous polynomial* of degree $d$ and denoted by $C^d_{\mathbf{k}}(\psi_1, ..., \psi_n)$ with $\mathbf{k}$ being a collection of internal spins.

*Remark.* The above definition appoints a way to construct multi-homogeneous polynomials. It has two steps: (*a*) fusing/coupling each input repeatedly. Then (*b*) fusing the processed vectors together (without repetition). As a result of spin recoupling theory, any complicated multi-homogeneous polynomial can be written as a linear combination of these standardized forms with 6j-symbols. In the following context, when we talk about multi-homogeneous polynomials, we always refer the ones described in Definition 2.3. For further use, we would also employ the notation $(d_1, ..., d_n) \vdash d$: $(d_1, ..., d_n)$ is called a *composition* of $d$ as $d_1 + \cdots + d_n = d$.

As an immediate result from Gordan theorem 2.2 and spin recoupling theory, we have:

**Theorem 2.4.** *Any SU(2)-equivariant polynomial $f : V_j^{\oplus n} \to W_{j'}$ can be written as a linear combination of standardized Clebsch-Gordan products $C^d_{\mathbf{k}}(\psi_1, ..., \psi_n)$.*

Note that in the above theorem, equivariant polynomials are defined on the direct sum of a single SU(2) irrep $V_j$. Even we can consider more general cases as in Theorem 2.2, the special case always happens in a real learning task, e.g., $n$ would represent the number of concerned atoms moving in the space and each of which is assigned a spin-$j$ vector standing for its dynamical/chemical states. To introduce permutation equivariance when relabeling these atoms, we make the following a simple observation:

**Lemma 2.5.** *Let $f : V_j^{\oplus n} \to W_{j'}$ be an $S_{n-1}$-invariant on its last $n - 1$ variables. Then the function $F$ defined by*

$$F(\psi_1, \psi_2, ..., \psi_n) = \big(f(\psi_1, \psi_2, ..., \psi_n), f(\psi_2, \psi_1, ..., \psi_n), ..., f(\psi_n, \psi_1..., \psi_{n-1})\big). \tag{5}$$

*is $S_n$-equivariant from $V_j^{\oplus n}$ to $W_{j'}^{\oplus n}$. Moreover, any $S_n$-equivariant function is written in this form.*

*Proof.* For first statement, let $\sigma \in S_n$, then by definition

$$F(\sigma \cdot (\psi_1, \psi_2, ..., \psi_n)) = \big(f(\psi_{\sigma^{-1}(1)}, \psi_{\sigma^{-1}(2)}, ..., \psi_{\sigma^{-1}(n)}), ..., f(\psi_{\sigma^{-1}(n)}, \psi_{\sigma^{-1}(1)}..., \psi_{\sigma^{-1}(n-1)})\big) \tag{6}$$

8

Since $f$ is invariant on its last $n-1$ variables,

$$f(\psi_{\sigma^{-1}(k)}, \psi_{\sigma^{-1}(2)}, ..., \psi_{\sigma^{-1}(n)}) = f(\psi_{\sigma^{-1}(k)}, \psi_1, ..., \psi_{\sigma^{-1}(k)-1}, \psi_{\sigma^{-1}(k)+1}, ..., \psi_n) \quad (7)$$

for $k = 1, ..., n$. Note that this identity should be revised when $\sigma^{-1}(k) = 1$ or $n$, but we omit these trivial details. On the other hand,

$$\sigma \cdot \big(F(\psi_1, \psi_2, ..., \psi_n)\big) = \sigma \cdot \big(f(\psi_1, \psi_2, ..., \psi_n), f(\psi_2, \psi_1, ..., \psi_n), ..., f(\psi_n, \psi_1..., \psi_{n-1})\big) \quad (8)$$

and permuted vectors can be written as Eq.(7), which establishes the $S_n$-equivariance.

Now let $F = (f_1, ..., f_n)$ be an arbitrary $S_n$-equivariant function. To prove the second part, we simply need to check the actions of transpositions $\sigma = (i, j) \in S_n$. For instance, acting on $F$ by $\sigma = (1, 2)$, we have

$$\begin{aligned}
F(\sigma \cdot (\psi_1, \psi_2, ..., \psi_n)) &= (f_1(\psi_2, \psi_1, ..., \psi_n), f_2(\psi_2, \psi_1, ..., \psi_n), ..., f_n(\psi_2, \psi_1, ..., \psi_n)) \\
&= (f_2(\psi_1, \psi_2, ..., \psi_n), f_1(\psi_1, \psi_2, ..., \psi_n), ..., f_n(\psi_1, \psi_2, ..., \psi_n)) \\
&= (\sigma \cdot F)(\psi_1, \psi_2, ..., \psi_n).
\end{aligned} \quad (9)$$

This indicates that

$$f_1(\psi_2, \psi_1, ...., \psi_n) = f_2(\psi_1, \psi_2, ...., \psi_n), \quad \forall \phi \in V_j. \quad (10)$$

Applying $(1, 3)$, we have

$$f_1(\psi_1, \psi_2, \psi_3, ..., \psi_n) = f_3(\psi_3, \psi_2, \psi_1, ..., \psi_n), \quad \forall \phi \in V_j. \quad (11)$$

Continuing by induction, we can conclude that

$$F(\psi_1, \psi_2, ..., \psi_n) = \big(f(\psi_1, \psi_2, ..., \psi_n), f(\psi_2, \psi_1, ..., \psi_n), ..., f(\psi_n, \psi_1..., \psi_{n-1})\big). \quad (12)$$

To verify that $f$ is $S_{n-1}$-invariant, we check Eq.(9) again which shows that $f_i(\psi_1, ...., \psi_n) = f(\psi_i, \psi_1, ..., \psi_n)$ is invariant when exchanging $\psi_1$ and $\psi_2$ for $i \geq 3$. Similar results hold for other transpositions and can be assembled to prove the invariance of $f$. □

*Remark.* In the language from [24]), we are talking about first-order $S_n$-equivariant function defined from $V_j^{\oplus n}$ to $W_{j'}^{\oplus n}$ where $S_n$ permutes input vectors. There are high order cases in which $S_n$ would permute tensor with more than one indices. However, higher order equivariance is generally not applicable here as $V_j, W_j$ are SU(2) irreps and do not carry a manifest $S_n$ action.

We now prove the main theorem:

**Theorem 2.6.** *A neural network equipped with fusion blocks can be used to build any $SU(2) \times S_n$ equivariant polynomial from $V_j^{\oplus n}$ to $W_{j'}^{\oplus n}$.*

*Proof.* With Algorithm 1 presented in the main text, we label atoms by $1, ..., n$ with atomic activations $(\psi_1, ..., \psi_n) \in V_j^{\oplus n}$. Updating with fusion blocks, we are able to create $SU(2) \times S_n$ equivariant polynomials $F : V_j^{\oplus n} \to W_j^{\oplus n}$ due to the employment of CG-products and aggregating at each layer. Any such function $F$ can be expanded as Eq.(5) by Lemma 2.5 in which each component $f(\psi_i, \psi_1, ..., \psi_n)$ is recorded with the $i$th atom being chosen as the center.

Since $f : V_j^{\oplus n} \to W_{j'}$ is both $S_{n-1}$-invariant on its last $n-1$ variables and $SU(2)$-equivariant, by Theorem 2.4,

$$f(\psi_1, ..., \psi_n) = \sum_{d, \mathbf{k}} c_{\mathbf{k}}^d C_{\mathbf{k}}^d(\psi_1, ..., \psi_n), \quad (13)$$

9

where $c_{\mathbf{k}}^d$ denotes the expansion coefficients. Furthermore, $f(\psi_1, ..., \psi_n)$ equals its symmetrization on the last $n-1$ variables:

$$\begin{aligned} f(\psi_1, ..., \psi_n) &= \frac{1}{(n-1)!} \sum_{\sigma \in S_{n-1}} f(\psi_1, \psi_{\sigma^{-1}(2)}, ..., \psi_{\sigma^{-1}(n)}) \\ &= \frac{1}{(n-1)!} \sum_{d, \mathbf{k}, \sigma \in S_{n-1}} c_{\mathbf{k}}^d C_{\mathbf{k}}^d(\psi_1, \psi_{\sigma^{-1}(2)}, ..., \psi_{\sigma^{-1}(n)}) \\ &= \sum_{\mathbf{k}} c_{\mathbf{k}}^d \overline{C}_{\mathbf{k}}^d(\psi_1, \psi_2, ..., \psi_n), \end{aligned} \qquad (14)$$

where we denote by $\overline{C}_{\mathbf{k}}^d(\psi_1, \psi_2, ..., \psi_n)$ the symmetrization of $C_{\mathbf{k}}^d(\psi_1, \psi_2, ..., \psi_n)$. To prove the universality, we just need to show that updating with fusion blocks produces symmetrized CG-products of the above kind, which can be done with the following procedure

1. For a given multi-degree $d = (d_1, ..., d_n)$, we first process $\psi_i$ by self-coupling (iterated CG-products) without aggregation as in Definition 2.3. We denote these pre-processed vectors by $C_{\mathbf{k}^j}^{d_j}(\psi_i)$ for $1 \leq i, j \leq n$.

2. We instantiate with the first atom being taken the center for reference. As an elementary example, assume $d = (d_1, d_2)$ with $d_i = 0$ for $i \geq 3$, then we apply the common CG-product, which corresponds to the simplest fusion diagram, on the central atom and each of its neighbor:

$$\sum_{i \geq 2} C^{k_{d_1}^1 k_{d_2}^2 J}(C_{\mathbf{k}^1}^{d_1}(\psi_1), C_{\mathbf{k}^2}^{d_2}(\psi_i)). \qquad (15)$$

Note that the summation appears now as we aggregate all center-neighbor interaction/coupling. It is exactly the symmetrization of standardized CG-products.

3. Assume $d_i = 0$ for $i \geq 4$, we update the central atom one more time:

$$\sum_{j \geq 2} C^{J' k_{d_3}^3 J}\left( \sum_{i \geq 2} \left( C^{k_{d_1}^1 k_{d_2}^2 J'}(C_{\mathbf{k}^1}^{d_1}(\psi_1), C_{\mathbf{k}^2}^{d_2}(\psi_i)) \right), C_{\mathbf{k}^3}^{d_3}(\psi_j) \right). \qquad (16)$$

General multi-homogeneous polynomials with required equivariance are constructed sequentially and we complete the proof. □

There is still one more subtle point need being clarify in the previous proof. It should be helpful to illustrate with the following example.

**Example.** Suppose $j = 0$, that is, atomic activations $x_1, ..., x_n$ are simply scalars and hence the SU(2)-equivariance holds trivially. We are left with the permutation equivariance. We still exemplify when the first atom is chosen as a center. Updating scalars with fusion blocks, we build polynomials as

$$M_\lambda(x_1, ..., x_n) = \sum_{i_1, ..., i_{n-1} \geq 2} x_1^{d_1} x_{i_1}^{d_2} \cdots x_{i_{n-1}}^{d_n}, \qquad (17)$$

where $\lambda$ denotes a composition $(d_1, ..., d_n)$ of the multi-degree $d$ (see the previous Remark). As $i_1, ..., i_{n-1}$ can be arbitrary numbers greater than one, there are cases when $i_r = i_s$ for $r \neq s$. On the other hand, there are **monomial symmetric polynomials** on the last $n-1$ variables:

$$m_\lambda(x_1, ..., x_n) := \sum_{\sigma \in S_{n-1}} x_1^{d_1} x_{i_{\sigma^{-1}(1)}}^{d_2} \cdots x_{i_{\sigma^{-1}(n-1)}}^{d_n}, \qquad (18)$$

10

which has no repetition on indices by definition. We note that

$$M_{\lambda=(d_1,d_2)} = m_{\lambda=(d_1,d_2)}, \quad M_{\lambda=(d_1,d_2,d_3)} = m_{\lambda=(d_1,d_2)} + m_{\lambda=(d_1,d_2,d_3)}, \cdots \cdots \quad (19)$$

It is well-known that monomial symmetric polynomials $m_\lambda$ can be used to build any permutation-invariant polynomial [8]. Even with repeated indices, Eq,19 shows that our polynomials $M_\lambda$ produced by fusion blocks also fulfill the task. This argument holds generally to the case when $j \neq 0$.

As a final step, we apply the so-called *Equivariant Stone-Weierstrass Theorem* which says that any continuous equivariant function can be approximated by equivariant polynomials [5] we conclude that:

**Corollary 2.7.** *The Cormorant architecture updated with fusion blocks permutation-equivariant and universal to approximate any continuous SO(3)-equivariant map.*

# 3 Architecture, hyper-parameter, and computational details

Here, we provide additional details on the architecture used, hyperparameter choices, and computational details for our experiments.

Cormorant over fusion diagram neural network was modified from the publically available cormorant package available at https://github.com/risilab/cormorant/, with the sole difference being the addition of the fusion block to the Atomlevel class in the network. To compensate for the additional cost of the network we changed the number of layers in the Cormorant_CG class to 2 and set the number of channes to 8 or the QM9 experiments and 16 for the MD17 experiments. (The number of channels was chosen primarily to ensure the model fit in the memory of the GPUs used.) The training was performed on a single NVIDIA V100 GPUs at single precision, at 256 Epochs each with a mini-batch size of 64. All other training hyperparameters were chosen to be identical to the ones used in [1]. This resulted in a total training time of 12 hours for the MD17 calculations and 20 hours for the QM9 calculations.

MACE over fusion diagram neural network was modified from the publically available MACE package available at https://github.com/ACEsuit/mace. While the MACE neural network was directly cloned for comparison. In non-adiabatic dynamics of stilbene simulations, we set the feature channel width of MoFD-sparse to 16 with $l = 0, 1, 2, 3$ without parity in the small-sample training, and we set the channel width to 32 in the training with more samples. For MACE, we set the feature channel width to 64 with $l = 0, 1, 2, 3$ without parity in the small-sample training, and we set the channel width to 128 in the training with more samples. For MoFD-dense, we set the feature channel width to 16 with $l = 0, 1, 2, 3$ without parity. In the training process of both MACE and MoFD, we set the maximal number of epochs to 500 with stopping patience of 30 epochs without improvement. The energy-to-force loss balance was enforced to 1:1000. The order of spherical harmonics is set to 3, while the size of the receptive field is set to 5 angstroms.

## 3.1 Electronic structure method

The fundamental electronic structure calculations in this study are performed using the Complete Active Space Self-Consistent Field (CASSCF) theory. CASSCF represents an advanced post-Hartree-Fock wavefunction-based method, which is particularly effective for systems with strong electron correlation effects. This method is distinguished by its approach to classifying

molecular orbitals into three distinct categories: inactive orbitals, active orbitals, and virtual orbitals.

Inactive orbitals are orbitals that are doubly occupied in all configurations and are thus excluded from the correlation treatment. The active orbitals form the 'active space' for the CASSCF calculations, which are a set of orbitals that can be occupied by a predefined number of electrons. The electrons in these orbitals are allowed to be distributed in all possible ways, leading to the generation of many electron configurations, thus capturing missing electron correlation effects in the Hartree-Fock calculation. Virtual orbitals are unoccupied in the reference state and lie above the active orbitals in energy. CASSCF iteratively optimizes both the molecular orbitals and the configuration interaction coefficients to achieve self-consistency.

CASSCF is particularly useful for studying excited states, transition states, and systems where traditional methods like Density Functional Theory (DFT) might fail due to strong electron correlation effects.

## 3.2 Dynamics details

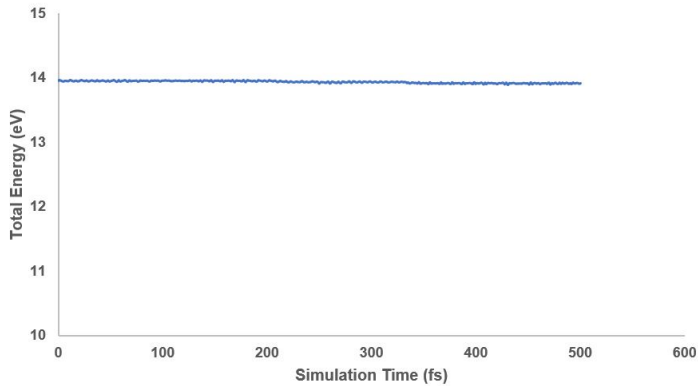

Figure 3: Illustration of conservation of energy over the dynamic simulation.

We used SHARC-MD package to run non-adiabatic dynamic simulations, the software is available at https://github.com/sharc-md/sharc. The compilation used SHARC-MD 2.1.2 with Intel One Compilers 2022 and Inter Math Kernel Library 2022. For SA-CASSCF calculations, OpenMolcas was used, which is available at https://gitlab.com/Molcas/OpenMolcas. The compilation used OpenMolcas 22.06 with Intel One Compilers 2022 and Inter Math Kernel Library 2022. Initial conformations are generated by Wigner-Sampling of the optimized ground-state structure with the same level of electronic structure method. For each conformation, a single-point calculation is performed to acquire the energy of states without spin-orbit calculations. To select initial excited-states, the MCH representation of the Hamiltonian is used to simulate delta-pulse excitation based on excitation energies and oscillators strengths with an excitation window of 0.0 to 10.0 eV. The threshold of total energy was set to 0.2 eV to account for inevitable active space rotation during the dynamic simulations. We initiated each trajectory with 0.5 fs as a single timestep and 500 fs as total length.

Table 1: Mean absolute error of various prediction targets on QM-9 (left) and conformational energies (in units of kcal/mol).

| Target | Unit | Cormorant | | Ours(CoFD) | |
|---|---|---|---|---|---|
| $\alpha$ | $a_0^3$ | 0.085 | (0.001) | 0.088 | (0.003) |
| $\Delta\epsilon$ | eV | 0.061 | (0.005) | 0.062 | (0.001) |
| $\epsilon_{\text{HOMO}}$ | eV | **0.034** | **(0.002)** | 0.0391 | (0.0008) |
| $\epsilon_{\text{LUMO}}$ | eV | 0.038 | (0.008) | 0.0347 | (0.0006) |
| $\mu$ | D | 0.038 | (0.009) | 0.035 | (0.001) |
| $C_v$ | $\frac{\text{cal}}{\text{mol K}}$ | **0.026** | **(0.000)** | 0.0272 | (0.0006) |
| $G$ | eV | 0.020 | (0.000) | **0.0135** | **(0.0002)** |
| $H$ | eV | 0.021 | (0.001) | **0.0132** | **(0.0004)** |
| $R^2$ | $a_0^2$ | 0.961 | (0.019) | **0.50** | **(0.02)** |
| $U$ | eV | 0.021 | (0.000) | **0.0130** | **(0.0004)** |
| $U_0$ | eV | 0.022 | (0.003) | **0.0133** | **(0.0003)** |
| ZPVE | meV | 2.027 | (0.042) | **1.43** | **(0.04)** |

Table 2: QM9 results on the Dimenet split. Benchmarks taken from [6].

| Target | Unit | SchNet | MGCN | DeepMoleNet | DimeNet | DimeNet[++] | Ours(CoFD) |
|---|---|---|---|---|---|---|---|
| $\mu$ | D | 0.0330 | 0.0560 | 0.0253 | 0.0286 | 0.0297 | 0.028 |
| $\alpha$ | $a_0^3$ | 0.235 | 0.0300 | 0.0681 | 0.0469 | 0.0435 | 0.0682 |
| $\epsilon_{\text{HOMO}}$ | meV | 41.0 | 42.1 | 23.9 | 27.8 | 24.6 | 37.8 |
| $\epsilon_{\text{LUMO}}$ | meV | 34.0 | 57.4 | 22.7 | 19.7 | 19.5 | 26.6 |
| $\Delta\epsilon$ | meV | 63.0 | 64.2 | 33.2 | 34.8 | 32.6 | 53.0 |
| $\langle R^2 \rangle$ | $a_0^2$ | 0.073 | 0.110 | 0.680 | 0.331 | 0.331 | 0.4128 |
| ZPVE | meV | 1.70 | 1.12 | 1.90 | 1.29 | 1.21 | 1.304 |
| $U_0$ | meV | 14.0 | 12.9 | 7.70 | 8.02 | 6.32 | 10.4 |
| $U$ | meV | 19.0 | 14.4 | 7.80 | 7.89 | 6.28 | 10.8 |
| $H$ | meV | 14.0 | 14.6 | 7.80 | 8.11 | 6.53 | 10.6 |
| $G$ | meV | 14.0 | 16.2 | 8.60 | 8.98 | 7.56 | 11.1 |
| $c_v$ | $\frac{\text{cal}}{\text{mol K}}$ | 0.0330 | 0.0380 | 0.0290 | 0.0249 | 0.0230 | 0.0242 |

## 4 Additional Numerical Results

Here, we give additional numerical results, complementing the ones given in the main text. First, give a more comprehensive treatment of our results on QM9. Whereas in the main text we gave results for QM9 on only a single target, in 1 we give the mean and standard deviation of our results over multiple replicates. Specifically, we give the mean and standard deviation over 3 replicates for $\Delta\epsilon$, $\epsilon_{\text{HOMO}}$, $\epsilon_{\text{LUMO}}$, $Cv$, $G$, $R^2$, $U$ and over 4 replicates for all other targets. To compare our work against other architectures we then trained this same model against the split of QM-9 given in the [7]; results are given in Table 2. We find that our model performs reasonably well, although it does not achieve state-of-the-art performance. However, we note that our ability to perform hyperparameter tuning was limited due to the expense of running the model and our comparative compute budget. We hope that subsequent hyperparameter tuning would improve the accuracy of our models.

In addition, we trained a variant of a model with the fusion block against the MD17 split introduced in [7]. This required making architectural changes compared with the original cormorant architecture: as implemented, the cormorant code evaluates radial functions on the displacements between an atom and itself. However, the original radial functions used are not

Table 3: Mean absolute testing error for the energies and forces for cormorant with a fusion block and various comparative models in the literature. Results taken from [2]

| Molecule | Target | Unit | SchNet | DimeNet | NequIP (l=3) | Ours |
|---|---|---|---|---|---|---|
| Aspirin | Energy | meV | 16.0 | 8.8 | **5.7** | 11.1 |
|  | Force | meV/Å$^2$ | 58.5 | 21.6 | **8.0** | 23.72 |
| Ethanol | Energy | meV | 3.5 | 2.8 | **2.2** | 2.6 |
|  | Force | meV/Å$^2$ | 16.9 | 10.0 | **3.1** | 9.4 |
| Malonaldehyde | Energy | meV | 5.6 | 4.5 | **3.3** | 4.7 |
|  | Force | meV/Å$^2$ | 28.6 | 16.6 | **5.6** | 20.4 |
| Naphthalene | Energy | meV | 6.9 | 5.3 | **4.9** | 8.1 |
|  | Force | meV/Å$^2$ | 25.2 | 9.3 | **1.7** | 21.4 |
| Salicylic Acid | Energy | meV | 8.7 | 5.8 | 4.6 | **4.2** |
|  | Force | meV/Å$^2$ | 36.9 | 16.2 | **3.9** | 8.9 |
| Toluene | Energy | meV | 5.2 | 4.4 | **4.0** | 6.0 |
|  | Force | meV/Å$^2$ | 24.7 | 9.4 | **2.0** | 16.6 |
| Uracil | Energy | meV | 6.1 | 5.0 | 4.5 | **3.7** |
|  | Force | meV/Å$^2$ | 24.3 | 13.1 | **3.3** | 6.9 |

differentiable at 0. Consequently, we instead used the radial functions

$$R(x) = \frac{x^2}{a} e^{-x^2/a} \tag{20}$$

where $a$ is a collection of learned parameters, initially evenly spaced on the interval $(0, 3]$. As before, each network was made of two neural network layers, this time with 64 channels for each layer. The network was trained on a linear combination of the mean square error in the energy and the force, with the total force error averaged over atoms and multiplied by a constant factor. The constant factor was chosen from 1, 10, and 100 by taking the value with the minimal validation error. Training proceeded on an NVIDIA A100 GPU for 2024 epochs, with a batch size of 64. We give the results in table 3.



\end{document}